\documentclass[11pt]{article}

\usepackage[preprint]{acl}

\usepackage{times}
\usepackage{latexsym}
\usepackage{multirow}

\usepackage[T1]{fontenc}

\usepackage[utf8]{inputenc}

\usepackage{microtype}

\usepackage{inconsolata}

\usepackage{graphicx}

\title{DependencyAI: Detecting AI Generated Text through Dependency Parsing}

\author{
  Sara Ahmed, Tracy Hammond \\
  Sketch Recognition Lab, Texas A\&M University \\
}

\begin{document}
\maketitle
\begin{abstract}
As large language models (LLMs) become increasingly prevalent, reliable methods for detecting AI-generated text are critical for mitigating potential risks. We introduce \textsc{DependencyAI}, a simple and interpretable approach for detecting AI-generated text using only the labels of linguistic dependency relations. Our method achieves competitive performance across monolingual, multi-generator, and multilingual settings. To increase interpretability, we analyze feature importance to reveal syntactic structures that distinguish AI-generated from human-written text. We also observe a systematic overprediction of certain models on unseen domains, suggesting that generator-specific writing styles may affect cross-domain generalization. Overall, our results demonstrate that dependency relations alone provide a robust signal for AI-generated text detection, establishing \textsc{DependencyAI} as a strong linguistically grounded, interpretable, and non-neural network baseline.

\end{abstract}

\section{Introduction}

Large Language Models (LLMs) are now widely used for content generation, automated assistance, and creative writing. While these systems are beneficial, they also raise concerns related to academic integrity and plagiarism \cite{cotton2024chatting, xie2023ai, chen2024plagiarism}, cybersecurity threats such as phishing and fake profiles \cite{roy2024chatbots, ayoobi2023looming}, and the large-scale spread of misinformation \cite{gao2024honestllm}. These risks underscore the need for reliable, interpretable, and robust methods for detecting AI-generated text.

In this work, we explore whether \emph{syntactic structure alone}, as represented by dependency relations, is sufficient to distinguish human-written from AI-generated text. We introduce \textsc{DependencyAI}, a lightweight detection method based exclusively on dependency relation sequences. Our approach (1) processes text using an off-the-shelf dependency parser, (2) represents dependency relation labels using TF--IDF n-grams, and (3) trains a LightGBM classifier to detect AI-generated text. By discarding lexical content and focusing only on dependency relation labels, we drastically reduce feature-space complexity while preserving structural information indicative of generative models’ writing patterns.

We evaluate our approach on the M4GT-Bench dataset across two tasks: (1) multi-way generation attribution under unseen domains and (2) multilingual detection across five languages. We also analyze feature importance to identify the dependency relations that most strongly contribute to detection, providing interpretability into the syntactic patterns associated with AI-generated text. We position \textsc{DependencyAI} not as a replacement for neural network detectors, but as an interpretable, linguistically grounded baseline that reveals structural regularities in AI-generated text.

\section{Related Work}

Detecting AI-generated text is commonly formulated as a binary classification task and can broadly be categorized into supervised, unsupervised, and watermarking approaches. Supervised approaches leverage annotated datasets to train models \cite{guo2023close, wang2024ai, antoun2023towards, abassy-etal-2024-llm, guo2024biscope}, and unsupervised methods rely on white-box features \cite{mitchell2023detectgpt, 10.5555/3692070.3692768, su2023detectllm, he2024mgtbench}. Watermarking approaches embed detectable signals during generation \cite{kirchenbauer2023watermark, 10.5555/3692070.3693308, 10.1609/aaai.v39i23.34684}. 

Comparative studies indicate that transformer models such as RoBERTa outperform feature-based pipelines, though often at the cost of interpretability \cite{ardeshirifar2025comparing}. Linguistically motivated approaches leverage lexical and structural properties of text to distinguish between human and machine authorship \cite{yadagiri2024detecting}. There are also hybrid methods combining transformer representations with linguistic features such as sentence length, part-of-speech distributions, and punctuation frequency \cite{pan-etal-2024-umuteam-semeval-2024}. Linguistic signals have also been studied in rewriting contexts; for example, through analyzing edit-distance patterns in AI-generated revisions \cite{mao2024raidar}.

Detection is further complicated in multi-generator, multi-domain, and multilingual settings, where generalization remains challenging \cite{wang2024m4, wang2024semeval, wang2024m4}. Prior work shows that fine-tuned multilingual detectors achieve the strongest overall performance, and linguistic similarity influences how well languages generalize to each other \cite{macko-etal-2023-multitude}. 

\section{Method}

Dependency parsing captures grammatical relationships between words, revealing the syntactic structure of a sentence. In this work, we discard lexical tokens and links entirely and retain only dependency relation labels (e.g., \texttt{nsubj}, \texttt{obj}, \texttt{ROOT}). Our pipeline consists of three steps. First, the text is processed using an off-the-shelf dependency parser. For English, we use SpaCy’s \texttt{en\_core\_web\_sm} model; for multilingual experiments, we use language-specific SpaCy models for Chinese, German, Italian, and Russian \cite{honnibal2020spacy}. Second, the resulting sequences of dependency labels are vectorized using TF--IDF with an \texttt{ngram\_range} of (1,2). Unigrams capture individual syntactic roles, while bigrams capture how these roles appear consecutively. Finally, we train a LightGBM classifier to distinguish between human-written and AI-generated text.

To improve interpretability, we analyze feature importance using the gain metric, which measures each feature’s contribution to reducing the loss function. This gives insight into the dependency relation patterns most strongly associated with detecting AI-generated text.

\begin{table}[t]
\centering
\small
\setlength{\tabcolsep}{3pt}
\begin{tabular}{llcccc}
\hline
\textbf{Detector} & \textbf{Test} & \textbf{Prec} & \textbf{Recall} & \textbf{F1} & \textbf{Acc} \\
\hline

\multirow{7}{*}{RoBERTa}
 & All        & 96.96 & 97.01 & \textbf{96.94}& \textbf{97.00}\\
 & arXiv      & 55.72 & 36.55 & 32.29 & 36.55 \\
 & PeerRead   & 70.58 & 70.12 & 66.89 & 69.47 \\
 & Reddit     & 77.21 & 74.49 & 71.66 & 74.49 \\
 & WikiHow    & 72.62 & 70.56 & 68.85 & 68.36 \\
 & Wikipedia  & 46.37 & 52.22 & 39.37 & 51.91 \\
 & OUTFOX     & 71.71 & 65.04 & 66.40 & 78.25 \\
\hline

\multirow{7}{*}{XLM-R}
 & All        & 90.73 & 90.37 & \textbf{90.16}& \textbf{90.17}\\
 & arXiv      & 51.29 & 43.88 & 41.71 & 43.88 \\
 & PeerRead   & 53.68 & 52.14 & 46.10 & 50.91 \\
 & Reddit     & 69.73 & 58.72 & 57.34 & 58.71 \\
 & WikiHow    & 65.73 & 60.84 & 58.45 & 57.38 \\
 & Wikipedia  & 60.04 & 42.55 & 38.80 & 41.95 \\
 & OUTFOX     & 51.94 & 42.44 & 43.00 & 52.10 \\
\hline

\multirow{7}{*}{GLTR-LR}
 & All        & 42.36 & 43.96 & 40.32 & 45.06 \\
 & arXiv      & 26.24 & 34.45 & 26.92 & 34.45 \\
 & PeerRead   & 42.20 & 44.10 & 39.04 & 44.32 \\
 & Reddit     & 45.54 & 46.28 & \textbf{41.70}& \textbf{46.28}\\
 & WikiHow    & 41.86 & 39.39 & 38.24 & 38.74 \\
 & Wikipedia  & 41.62 & 36.95 & 34.62 & 35.18 \\
 & OUTFOX     & 29.68 & 32.38 & 29.18 & 38.78 \\
\hline

\multirow{7}{*}{GLTR-SVM}
 & All        & 52.81 & 48.42 & 47.24 & 50.39 \\
 & arXiv      & 22.57 & 32.29 & 25.73 & 32.28 \\
 & PeerRead   & 34.10 & 39.10 & 34.81 & 39.40 \\
 & Reddit     & 43.21 & 46.19 & \textbf{40.94}& \textbf{46.19}\\
 & WikiHow    & 44.13 & 39.01 & 38.27 & 36.18 \\
 & Wikipedia  & 34.86 & 30.95 & 26.72 & 29.81 \\
 & OUTFOX     & 26.93 & 28.29 & 26.45 & 28.58 \\
\hline

\multirow{7}{*}{Stylistic-SVM}
 & All        & 78.95 & 37.10 & \textbf{47.31}& \textbf{35.26}\\
 & arXiv      & 44.89 & 8.52  & 12.71 & 8.27 \\
 & PeerRead   & 50.72 & 21.96 & 25.43 & 20.44 \\
 & Reddit     & 60.98 & 27.25 & 31.16 & 24.43 \\
 & WikiHow    & 56.23 & 34.57 & 37.04 & 26.71 \\
 & Wikipedia  & 47.94 & 21.21 & 27.77 & 16.06 \\
 & OUTFOX     & 48.28 & 27.46 & 32.50 & 26.80 \\
\hline

\multirow{7}{*}{NELA-SVM}
 & All        & 64.50 & 23.91 & \textbf{30.54}& \textbf{22.76}\\
 & arXiv      & 47.35 & 11.53 & 16.20 & 10.94 \\
 & PeerRead   & 44.63 & 20.00 & 20.97 & 18.72 \\
 & Reddit     & 42.77 & 24.27 & 27.87 & 20.72 \\
 & WikiHow    & 48.33 & 25.81 & 25.51 & 21.83 \\
 & Wikipedia  & 46.38 & 20.74 & 25.06 & 18.76 \\
 & OUTFOX     & 35.05 & 17.44 & 18.48 & 19.18 \\
\hline

\multirow{7}{*}{\textbf{DependencyAI}}
 & All        & 89.15& 88.80& \textbf{88.85}& \textbf{88.94}\\
 & arXiv      & 46.88& 28.52& 25.56& 28.53\\
 & PeerRead   & 62.68& 61.26& 57.51& 61.24\\
 & Reddit     & 67.46& 60.87& 57.48& 60.87\\
 & WikiHow    & 61.28& 49.27& 43.80& 44.30\\
 & Wikipedia  & 44.45& 34.03& 28.90& 34.30\\
 & OUTFOX     & 54.60& 49.42& 45.26& 59.08\\
\hline

\end{tabular}
\caption{Results for multi-generator detection in which classifiers are trained on the data of all domains except for the test domain. Comparison scores taken from \cite{wang2024m4gt}}
\label{tab:domain_generalization}
\end{table}

\section{Experimental Setup}

\begin{figure*}[t]
    \centering
    \includegraphics[width=1\textwidth]{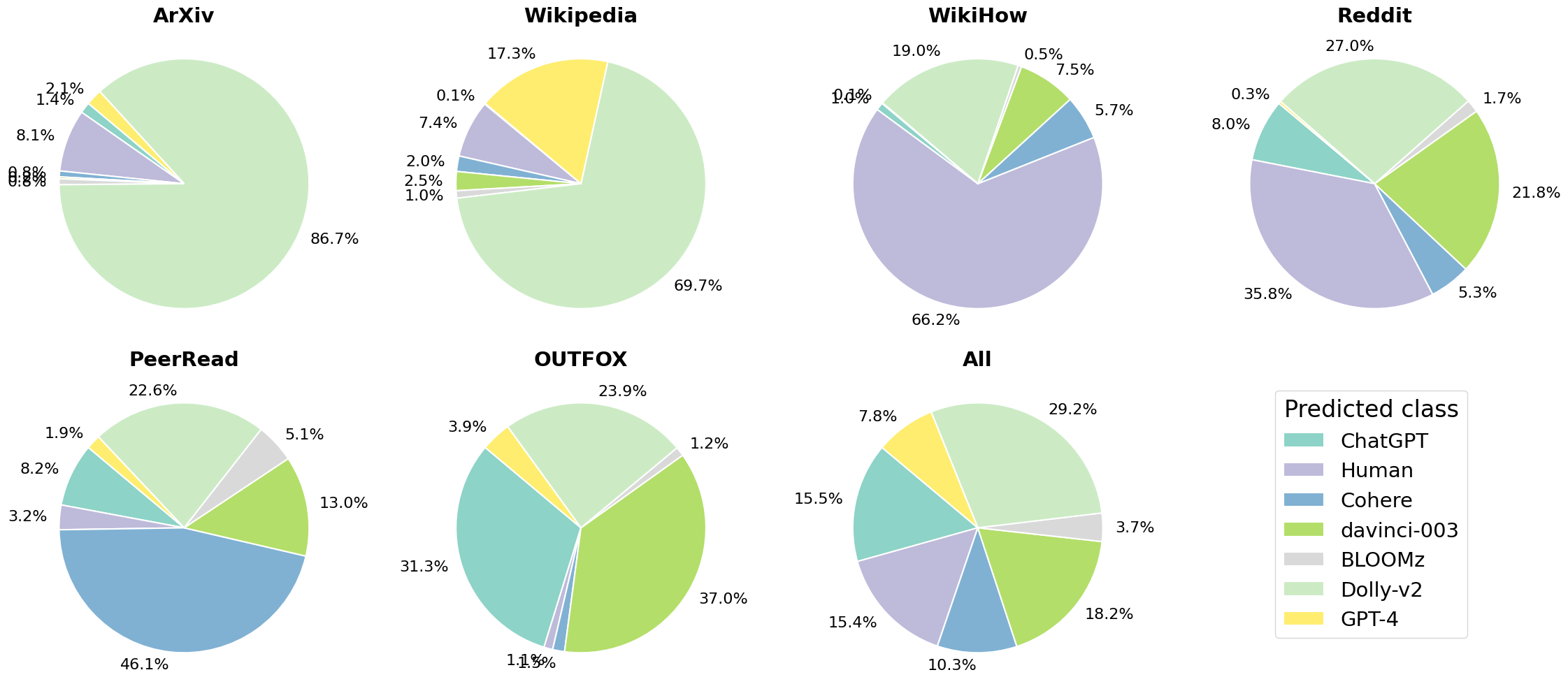}
    \caption{Distribution of prediction errors by predicted class for the multi-way detection task on domains excluded from training. Percentages indicate the proportion of errors attributed to each predicted class.}
    \label{fig:charts}
\end{figure*}

\subsection{Dataset}
We evaluate our approach on the M4GT-Bench dataset, a multilingual, multi-domain, and multi-generator benchmark for AI-generated text detection \cite{wang2024m4gt}. It consists of human-written text and outputs from multiple LLMs across diverse domains and languages. For the multilingual task, we use a subset of the multilingual portion, corresponding to languages supported by spaCy’s dependency parser; consequently, our results depend on both the parser’s language coverage and its parsing quality.

\subsection{Tasks}
We test our method on the following two tasks:
\begin{itemize}

    \item \textbf{Multi-way Detection}:  The goal is to identify human-written text and outputs from multiple generators while evaluating the classifier's ability to generalize to an unseen domain. Task taken from \cite{wang2024m4gt}.

    \item \textbf{Multilingual Detection}: The goal is to distinguish human-written from AI-generated text across Chinese, English, German, Italian, Russian, and the aggregated dataset (“All”).
\end{itemize}

\subsection{Baselines}
We compare \textsc{DependencyAI} against feature-based detectors (GLTR \cite{Gehrmann2019GLTR}, stylistic \cite{li2014authorship}, and NELA features \cite{horne2019robust}) as well as pretrained neural network models, RoBERTa \cite{liu2019roberta} and XLM-R \cite{conneau2020unsupervised}. We include the scores reported by \cite{wang2024m4gt} for comparison. For the multilingual task, we fine-tune an XLM-R model as our baseline.

\section{Results}

\subsection{Multi-way Detection}

Table 1 summarizes our results for multi-class classification across seven labels: human, davinci-003, ChatGPT, GPT-4, Cohere, Dolly-v2, and BLOOMz, with evaluation focused on generalization to a domain excluded during training. Our method achieves the strongest performance among all feature-based approaches, with an accuracy of 88.85 and an F1 score of 88.94. It surpasses all feature-based accuracy baselines, NELA-SVM, Stylistic-SVM, and GLTR, except for GLTR on ArXiv and GLTR-LR on Wikipedia. Notably, our method also exceeds XLM-R on PeerRead, Reddit, and OUTFOX. In line with findings from \cite{wang2024m4gt}, our class-wise analysis shows that davinci-003 remains the most difficult generator to detect, whereas BLOOMz is the easiest. 

Figure 1 shows prediction errors by predicted class across domains, revealing clear domain-specific confusion patterns. These patterns suggest that certain syntactic features may be associated with specific generation models and affect cross-domain generalization through overprediction. For example, in ArXiv, the majority of errors (86.7\%) stem from misclassification as Dolly-v2, and a similar trend appears in Wikipedia, where 69.7\% of errors are also attributed to Dolly-v2—both domains characterized by academic or informational writing. In contrast, WikiHow and Reddit, domains reflecting instructional and forum-based writing styles, show the highest proportion of errors misclassified as human text (66.2\% and 35.8\%, respectively). In PeerRead, a dataset of peer reviews, most misclassifications are attributed to Cohere (46.1\%) and in OUTFOX, which consists of student essays, most misclassifications are from davinci-003 (37.1\%) and ChatGPT (31.3\%). The “All” category, which trains on random data from all generators, has the most diverse distribution of prediction errors.

\subsection{Multilingual Detection}
Results for multilingual detection are shown on Table 2. \textsc{DependencyAI} achieves the strongest overall performance, surpassing XLM-R across most languages, specifically English, German, Russian, and the multilingual "All" category. Performance is particularly strong for Russian, where our method achieves 99.50 F1 and accuracy. These results suggest that dependency relations generalize well across languages despite using no lexical information.

\subsection{Effect of Dependency Relation N-grams}
Figure 2 examines the effect of n-gram size on detection performance. An n-gram range of (1,2) yields the largest performance gain, whereas extending it to (1,3) results in only marginal improvements for English and Chinese, slight declines for Italian, German, and the overall dataset, and largely stable performance for Russian. The most substantial improvement is observed for Chinese, where expanding from unigrams to bigrams increases accuracy by approximately five percentage points.

\begin{table}[t]
\centering
\small
\setlength{\tabcolsep}{4pt} 
\begin{tabular}{llcccc}
\hline
\textbf{Detector} & \textbf{Test} & \textbf{Prec} & \textbf{Recall} & \textbf{F1} & \textbf{Acc} \\
\hline

\multirow{6}{*}{XLM-R} 
 & All     & 86.64 & 99.49 & 92.62 & 91.80 \\
 & Chinese & 92.71 & 99.66 & \textbf{96.06}& \textbf{95.93}\\
 & English & 88.72 & 99.36 & 93.74 & 93.09 \\
 & German  & 78.85 & 99.57 & 88.01 & 86.43 \\
 & Italian & 98.86 & 1.00  & \textbf{99.43}& \textbf{99.42}\\
 & Russian & 96.06 & 97.50 & 96.77 & 96.75 \\
\hline

\multirow{6}{*}{\textbf{DependencyAI}}& All     & 96.02& 96.21& \textbf{96.12}& \textbf{95.98}\\
 & Chinese & 93.48& 91.74& 92.60& 92.71\\
 & English & 98.02& 98.05 & \textbf{98.04}& \textbf{97.96}\\
 & German  & 98.27& 97.57 & \textbf{97.92}& \textbf{97.93}\\
 & Italian & 96.42& 97.37& 96.89& 96.87\\
 & Russian & 99.50& 99.50& \textbf{99.50} & \textbf{99.50} \\
\hline

\end{tabular}
\caption{Multilingual evaluation of detectors trained and tested on the same language. Bold indicates the best F1 and Accuracy scores for each language.}
\label{tab:detector_column}
\end{table}

\begin{figure}
    \centering
    \includegraphics[width=1\linewidth]{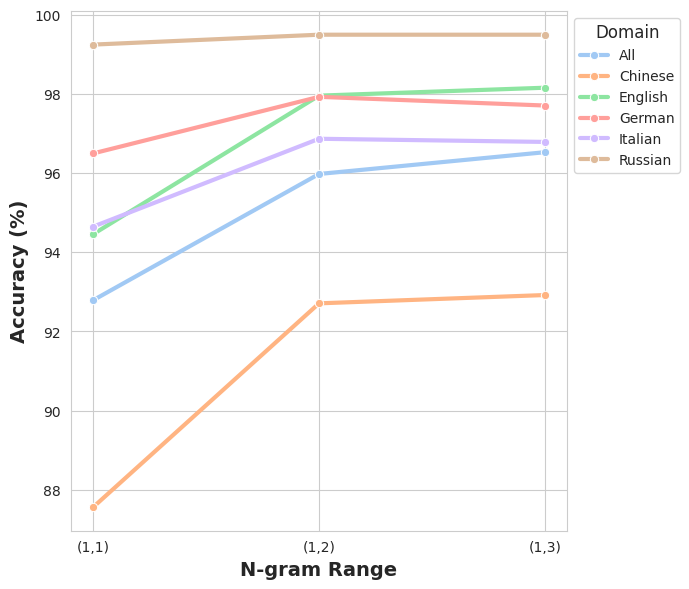}
    \caption{Dependency relation n-gram range vs. accuracy by language}
    \label{fig:ngrams}
\end{figure}

\begin{figure*}[t]
    \centering
    \includegraphics[width=\textwidth]{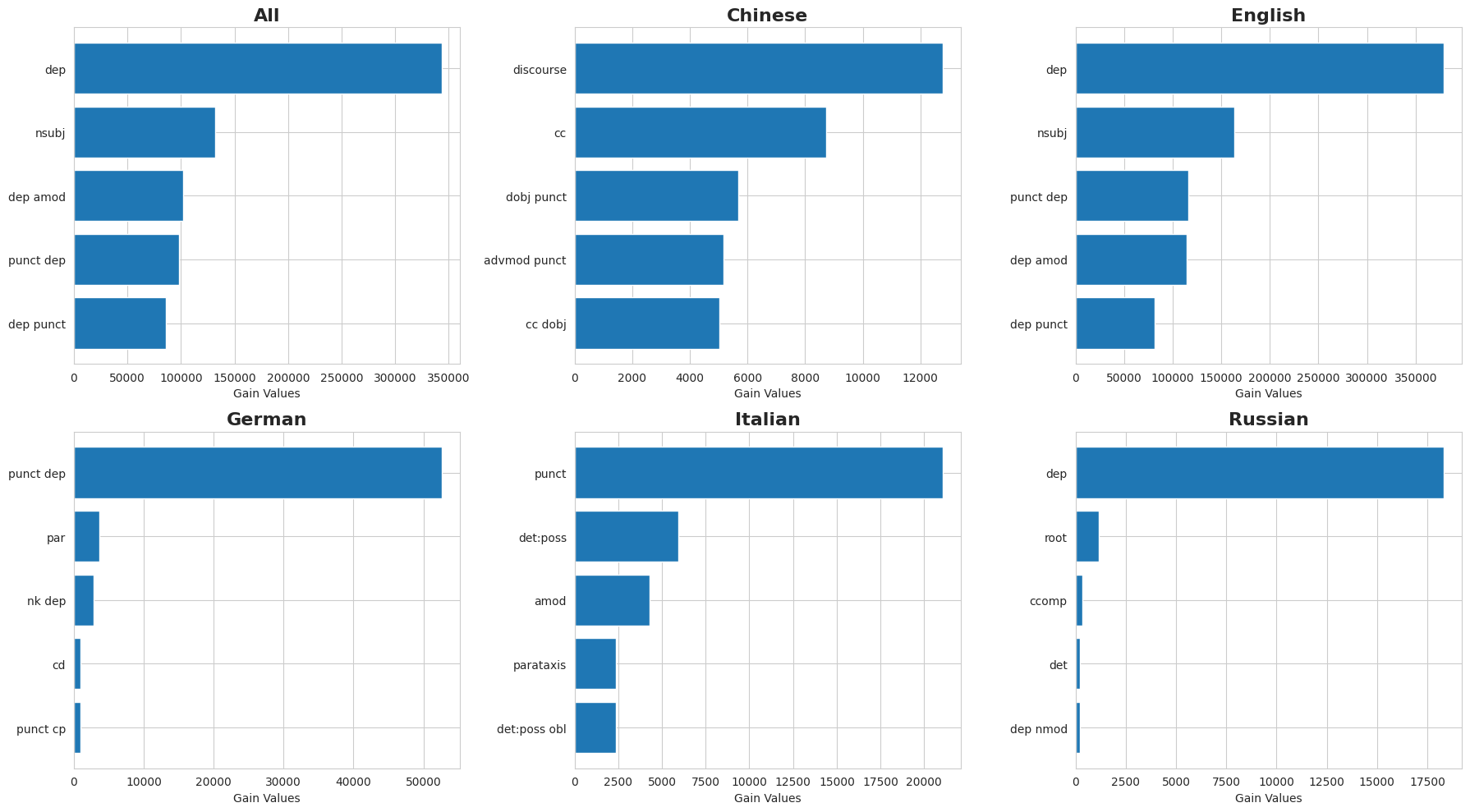}
    \caption{Top-5 feature importance by gain across languages. Each subplot corresponds to a language (Chinese, English, German, Italian, Russian) and the aggregated dataset (All), showing the five dependency features with the highest gain values for each.}
    \label{fig:multilingual_features}
\end{figure*}

\section{Feature Importance}

Figure 3 shows the top five most influential dependency relations for AI-generated text detection in each language, ranked by gain-based feature importance. In English, the highest-gain features are dominated by the unspecified dependency relation (\textit{dep}) and its n-grams, including punctuation–unspecified dependency (\textit{punct dep}), unspecified dependency–adjectival modifier (\textit{dep amod}), and unspecified dependency–punctuation (\textit{dep punct}). The unspecified dependency relation may arise from unusual grammatical constructions or limitations in the parsing software. The second most important feature is the placement of the nominal subject (\textit{nsubj}). Overall, this distribution of high-gain dependency relations indicates that unspecified dependencies—and their relation to punctuation and adjectival modifiers—along with nominal subject placement, provide the strongest signals for distinguishing between human-written and AI-generated English text.

Chinese gives weight to \textit{discourse} markers, and then coordinating conjunctions (\textit{cc}). Other influential features include direct object–punctuation ngram (\textit{dobj punct)}, adverbial modifier–punctuation ngram (\textit{advmod punct}), and coordinating conjunction–direct object ngram (\textit{cc dobj}). The prominence of these features suggests that the placement and interaction of discourse markers, coordinating conjunctions, direct objects, and punctuation play a central role in differentiating Chinese AI-generated and human written text.

For Italian, punctuation (\textit{punct}) emerges as the most salient feature, followed by possessive determiners (\textit{det:poss}). Adjectival modifiers (\textit{amod}) also play a strong role, alongside \textit{parataxis}, a structure in which clauses are placed together without explicit syntactic linking. Additionally, the possessive determiner–oblique nominal ngram pattern (\textit{det:poss obl}) shows a notable influence. Oblique nominals mark nouns or noun phrases that are not core arguments and typically convey supplementary or peripheral information. The prominence of these features suggests characteristic patterns of sentence linking and information structuring that is indicative of differences between Italian AI-generated and human-written text.

German shows a strong emphasis on punctuation-related interactions, particularly involving unspecified dependencies (\textit{punct dep}) and complementizers (\textit{punct cp}), and parenthetical elements (\textit{par}). Noun kernel elements (\textit{nk}), which mark constituents belonging to the core of the noun phrase, as well as coordinating conjunctions (\textit{cd}), also emerge as important features. Overall, the prominence of unspecified dependencies, punctuation usage, noun phrase structure, and coordination provides valuable signals for distinguishing between AI-generated and human-written German text.

Russian differs from the other languages in not emphasizing punctuation as a key marker. Similar to English, unspecified dependencies (\textit{dep}) carry the highest, substantial gain, later followed by the n-gram combining unspecified dependency and nominal modifier (\textit{dep nmod}). Additionally, the position of the sentence \textit{root}, clausal complements (\textit{ccomp}), and determiners (\textit{det}) also provide informative cues for differentiating Russian AI-generated from human-written text.

\section{Conclusion}

We introduced \textsc{DependencyAI} as a strong, interpretable, and linguistically grounded baseline. By discarding lexical content and relying on only dependency relation sequences, our method isolates structural patterns that distinguish human-written and machine-generated text. Across multilingual and multi-generator evaluations on M4GT-Bench, \textsc{DependencyAI} often exceeds traditional feature-based baselines and, in several cases, XLM-R. In multi-way detection with multiple generators, we observe systematic overprediction of certain models when evaluated on unseen domains, which may be influenced by writing styles.

Beyond empirical performance, a key contribution of this work lies in its interpretability. Gain-based feature analysis reveals that, depending on the language, certain syntactic features are key to distinguishing AI-generated text. Our findings demonstrate that syntactic structure constitutes a robust signal for AI-generated text detection.

\section{Limitations}

Despite its strengths, this work has some limitations. First, \textsc{DependencyAI} depends on the quality and availability of dependency parsers, which may degrade for noisy text, informal writing, or low-resource languages. Second, our experiments do not consider adversarial settings such as human-in-the-loop rewriting or post-editing; evaluating robustness under such conditions remains an important direction for future work. Finally, this approach is not applicable to code generation, as LLM-generated code cannot be detected using these linguistic features.

\bibliography{custom}
\end{document}